\documentclass{uncecomp2017}

\usepackage{url}
\usepackage{graphicx}
\usepackage{amsmath,amssymb}
\usepackage{amsthm}
\usepackage{times}
\usepackage{framed}
\usepackage{svg}
\usepackage{breqn} 
\usepackage{authblk}	

\newcommand{\ee}{\end{equation}}
\newcommand{\be}{\begin{equation}}
\newcommand{\ec}{\end{center}}
\newcommand{\bc}{\begin{center}}
\newcommand{\eea}{\end{eqnarray}}
\newcommand{\bea}{\begin{eqnarray}}
\newcommand{\bd}{\begin{description}}
\newcommand{\ed}{\end{description}}
\newcommand{\bi}{\begin{itemize}}
\newcommand{\ei}{\end{itemize}}
\newcommand{\pa}{\partial}
\newcommand{\bs}{\boldsymbol}

\newcommand{\tm}{\textrm}
\newcommand{\tit}{\textit}

\newcommand{\tbf}{\textbf}
\newcommand{\emat}{\end{pmatrix}}
\newcommand{\bmat}{\begin{pmatrix}}
\newcommand{\esmat}{\end{smallmatrix}\right)}
\newcommand{\bsmat}{\left(\begin{smallmatrix}}
\newcommand{\bes}{\begin{equation}\begin{split}}
\newcommand{\ees}{\end{split}\end{equation}}

\newcommand{\btc}{\bs{\theta}_c}

\newcommand{\refeq}[1]{Equation (\ref{#1})}

\title{Probabilistic reduced-order modeling for stochastic partial differential equations}

\author{Constantin Grigo$^1$ and Phaedon-Stelios Koutsourelakis$^1$}

\heading{Constantin Grigo and Phaedon-Stelios Koutsourelakis}

\address{$^1$Continuum Mechanics Group,\\
Department of Mechanical Engineering,\\
Technical University of Munich,\\
D-85748 Garching b. M\"unchen}

\keywords{Reduced-order modeling, generative Bayesian model, SPDE, effective material properties}

\abstract{We discuss a Bayesian formulation to coarse-graining (CG) of PDEs where the coefficients (e.g. material
parameters) exhibit random, fine scale variability. The direct solution to such problems requires grids that are
small enough to resolve this fine scale variability which unavoidably requires the repeated solution of very 
large systems of algebraic
equations. 

We  establish  a  physically  inspired,  data-driven  coarse-grained  model  which
learns a low-dimensional set of microstructural features that are \tit{predictive} of the fine-grained model
(FG) response. Once learned, those features provide a sharp distribution over the coarse scale effective
coefficients of the PDE that are most suitable for prediction of the fine scale model output. 

This ultimately allows to replace the computationally expensive FG by a generative probabilistic model based on 
evaluating the
much cheaper CG several times. Sparsity enforcing priors further increase predictive efficiency  and  reveal  
microstructural 
 features that are important in predicting the FG response. Moreover, the model yields 
probabilistic rather
than single-point predictions, which enables the quantification of the unavoidable epistemic uncertainty that is present
due to the information loss that occurs during the coarse-graining process.}

\begin{document}

\section{INTRODUCTION}

Many engineering design problems such as  flow in porous media and mechanical properties of composite materials
require simulations that are capable of resolving the  microstructure of the underlying medium. If the material
components under consideration exhibit fine-scale heterogeneity, popular discretization schemes (e.g. finite elements)
yield very large systems of algebraic equations. Pertinent solution strategies at best (e.g. multigrid methods) scale
linearly with the dimension of the unknown state vector.  Despite the ongoing improvements in computer hardware,
repeated solutions of such problems, as is required in the context of uncertainty quantification (UQ), poses 
insurmountable
difficulties.

It is obvious that viable strategies for such problems, as  well as a host of other deterministic problems where 
repeated
evaluations are needed such as inverse, control/design problems etc, should focus on methods that exhibit
\tit{sublinear} complexity with respect to the dimension of the original problem. In the context of UQ a popular and
general such strategy involves the use of surrogate models or emulators which attempt to learn the input-output map
implied by the fine-grained model.  Such models, e.g. Gaussian Processes \cite{Williams2005a}, polynomial chaos
expansions \cite{Gahem1991}, neural nets \cite{Bishop1995} and many more, are trained  on a finite set of fine-grained
model runs. Nevertheless, their performance is  seriously impeded by the curse of dimensionality, i.e. they usually
become inaccurate for input dimensions larger than a few tens or hundreds,   or equivalently, the number of FG runs
required to achieve an acceptable level of accuracy grows exponentially fast with the input dimension.

Alternative  strategies for high-dimensional problems make use of  multi-fidelity models \cite{Kennedy2000a,
Perdikaris2015} as inexpensive predictors of the FG output. As shown in \cite{Koutsourelakis2009}, lower-fidelity 
models
whose output deviates significantly from that of the FG can still yield accurate estimates with significant
computational savings, as long as the outputs of the models exhibit statistical dependence. In the case of  PDEs where 
finite elements are employed as the FG, multi-fidelity solvers can be simply obtained by using coarser 
discretizations in space or time. While linear and nonlinear dimensionality reduction techniques are suitable 
for  dealing with high-dimensional inputs \cite{ma_kernel_2011}, it is known which of the microstructural 
features are actually predictive of   FG outputs \cite{DBLP:journals/corr/physics-0004057}.

The model proposed in the present paper attempts to address this question. By using a two-component Bayesian network, we
are able to predict fine-grained model outputs based on only a finite number of training data runs and a repeated solution
of a much coarser model. Uncertainties can be easily quantified as our model leads to probabilistic rather than
point-like predictions.

\section{THE FINE-GRAINED MODEL}
\label{sec:SPDE}
Let $\left(\Omega, \mathcal{F}, P\right)$ be a probability space. Let $\mathcal H$ be the Hilbert space of functions
defined over the domain $\mathcal D$ over which the physical problem is defined. We consider problems in the context of 
heterogeneous media which 
exhibit properties given by a random process $\lambda(\bs x, \xi)$ defined over the product space $\mathcal D \times
\Omega$. The corresponding stochastic PDE may be written as
\be
\mathcal L\left(\bs x, \lambda(\bs x, \xi)) u(\bs x, \lambda(\bs x, \xi)\right) = f(\bs 
x), \qquad \tm{+B.C.}
\label{SPDE}
\ee
where $\mathcal L$ is a stochastic differential operator and $\bs x \in \mathcal D$, $\xi \in \Omega$ are elements of
the physical domain and the sample space, respectively. Discretization of the random process \\$\lambda(\bs x, \xi)
\underbrace{\longrightarrow}_{\textrm{discretize}}  \bs \lambda_f \in \mathbb{R}^{n_{\bs\lambda_f}}$ as well as the
governing equation leads to a system of $n_f$ (potentially nonlinear) algebraic equations, which can be written in
residual form as
\be
\bs{r}_f(\bs U_f; \bs \lambda_f) = \bs 0,
\label{FG}
\ee
where $\bs U_f(\bs \lambda_f) \in \mathbb{R}^{n_f}$ is the $n_f$-dimensional discretized solution vector for a given
$\bs \lambda_f$ and \newline $\bs{r}_f: \mathbb{R}^{n_f} \times \mathbb{R}^{n_{\bs \lambda_f}} \to  \mathbb{R}^{n_f}$
the discretized residual vector. It is the model  described by equation \eqref{FG} which is denoted as the fine-grained
model (FG) henceforth.

\section{A GENERATIVE BAYESIAN SURROGATE MODEL}
Let
\be
p(\bs \lambda_f) = \int \delta(\bs \lambda_f - \bs \lambda_f(\xi)) p(\xi)d\xi
\label{eq:inputpdf}
\ee
be the density of $\bs \lambda_f$. The density of the fine-scale response $\bs U_f$ is then given by
\be
p(\bs U_f) = \int p(\bs U_f|\bs \lambda_f)p(\bs \lambda_f)d\bs \lambda_f,
\ee
where the conditional density $p(\bs U_f|\bs \lambda_f)$ degenerates to a $ \delta(\bs U_f - \bs U_f(\bs \lambda_f))$
when the only uncertainties in the problem are due to $\bs{\lambda}_f$.

The objective of this paper is to approximate this input-output map implied by $\bs U_f(\bs \lambda_f)$, or
equivalently in terms of probability densities, the  conditional distribution $p(\bs U_f|\bs \lambda_f)$. The latter
case can also account for problems where additional sources of uncertainty are present and the input-output map is 
stochastic. To that end, we introduce a coarse-grained model (CG) leading to an approximate  distribution $\bar{p}(\bs
U_f|\bs \lambda_f)$ which will be trained on  a limited number of FG solutions $\mathcal D_N = \left\{\bs
\lambda_f^{(i)}, \bs U_f^{(i)}(\bs \lambda_f^{(i)})\right\}_{i = 1}^{N}$.

Our approximate model $\bar{p}(\bs U_f|\bs \lambda_f)$ employs a set of latent \cite{Bishop1998},
reduced, collective variables which we denote by  $\bs \lambda_c \in \mathbb{R}^{n_{\bs \lambda_c}}$ for reasons that
will be apparent in the sequel, such that
\be
\begin{split}
\bar{p}(\bs U_f|\bs \lambda_f) & = \int \bar{p}(\bs U_f, \bs \lambda_c |\bs \lambda_f)d\bs \lambda_c \\
&= \int \underbrace{ \bar{p}(\bs U_f | \bs \lambda_c)}_{\tm{decoder}} \underbrace{\bar{p}(\bs \lambda_c 
|\bs \lambda_f)}_{\tm{encoder}}d\bs \lambda_c.
\end{split}
\label{eq:autoencoder}
\ee
As it can be understood from the equation above, the latent variables $\bs \lambda_c$ act as  a probabilistic
filter (encoder) on the FG input $\bs \lambda_f$, retaining the features necessary for predicting the FG output $\bs
U_f$.
In order for $\bar{p}(\bs U_f|\bs \lambda_f)$ to approximate well  ${p}(\bs U_f|\bs \lambda_f)$, the 
latent variables $\bs \lambda_c$ should not
simply be the outcome of a dimensionality reduction on $\bs \lambda_f$. Even if  $\bs \lambda_f$ is amenable to such  a 
dimensionality
reduction, it is not necessary that the $\bs \lambda_c$ found would be predictive of $\bs U_f$. Posed differently, it is
not important that $\bs \lambda_c$ provides a high-fidelity encoding of $\bs \lambda_f$ but it suffices that it is
capable of providing a good prediction of the corresponding $\bs U_f(\bs \lambda_f)$.

The aforementioned desiderata do not unambiguously define the form of the encoding/ decoding densities in
\eqref{eq:autoencoder} nor the type/dimension of the latent variables $\bs \lambda_c$. In order to retain some of the
physical and mathematical structure of the FG, we propose employing a coarsened, discretized version of the original
continuous equation \eqref{SPDE}. In residual form this can again be written as
\be
\bs{r}_{c}(\bs U_c; \bs \lambda_c) = \bs{0},
\label{CG}
\ee
where $\bs U_c \in \mathbb{R}^{n_c}$ is the $n_c$-dimensional ($n_c \ll n_f$) discretized  solution and $\bs{r}_c:
\mathbb{R}^{n_f} \times \mathbb{R}^{n_{\bs \lambda_c}} \to \mathbb{R}^{n_c}$ is the discretized residual vector. Due to
the significant discrepancy in dimensions $n_c \ll n_f$, the cost of solving the CG in \eqref{CG} is negligible
compared to the FG in \eqref{FG}.


It is clear that $\bs \lambda_c$ plays the role of effective/equivalent properties but it is not obvious (except for
some limiting cases where homogenization results can be invoked \cite{Torquato2001}) how these should depend on the
fine-scale input $\bs \lambda_f$ nor how the solution $\bs U_c(\bs \lambda_c)$ of the CG should relate to $\bs U_f(\bs
\lambda_f)$ in \eqref{FG}. Furthermore, it is important to recognize a priori that the use of the reduced variables 
$\bs \lambda_c$ in combination with  the coarse model in \eqref{CG} would in general imply some information loss. The
latter should introduce an additional source of uncertainty in the predictions of the fine-scale output $\bs U_f$
\cite{Weinan2011}, even in the limit of infinite training data. For that purpose and in agreement with
\eqref{eq:autoencoder}, we propose a generative probabilistic  model composed of the following two densities:
\bi
\item A probabilistic mapping from $\bs \lambda_f$ to $\bs \lambda_c$, which determines 
the effective properties $\bs \lambda_c$ given $\bs \lambda_f$. We write 
this as $p_c(\bs \lambda_c|\bs \lambda_f,
\bs\theta_c)$ where  $\bs \theta_c$ denotes a set of model parameters,
\item A coarse-to-fine map $p_{cf}(\bs U_f|\bs U_c, \bs \theta_{cf})$, which is the PDF 
of the FG output $\bs U_f$
given the output $\bs U_c$ of the CG. It is parametrized by $\bs \theta_{cf}$.
\ei
\begin{figure}[t]
\begin{center}
\includegraphics[width=.5\textwidth]{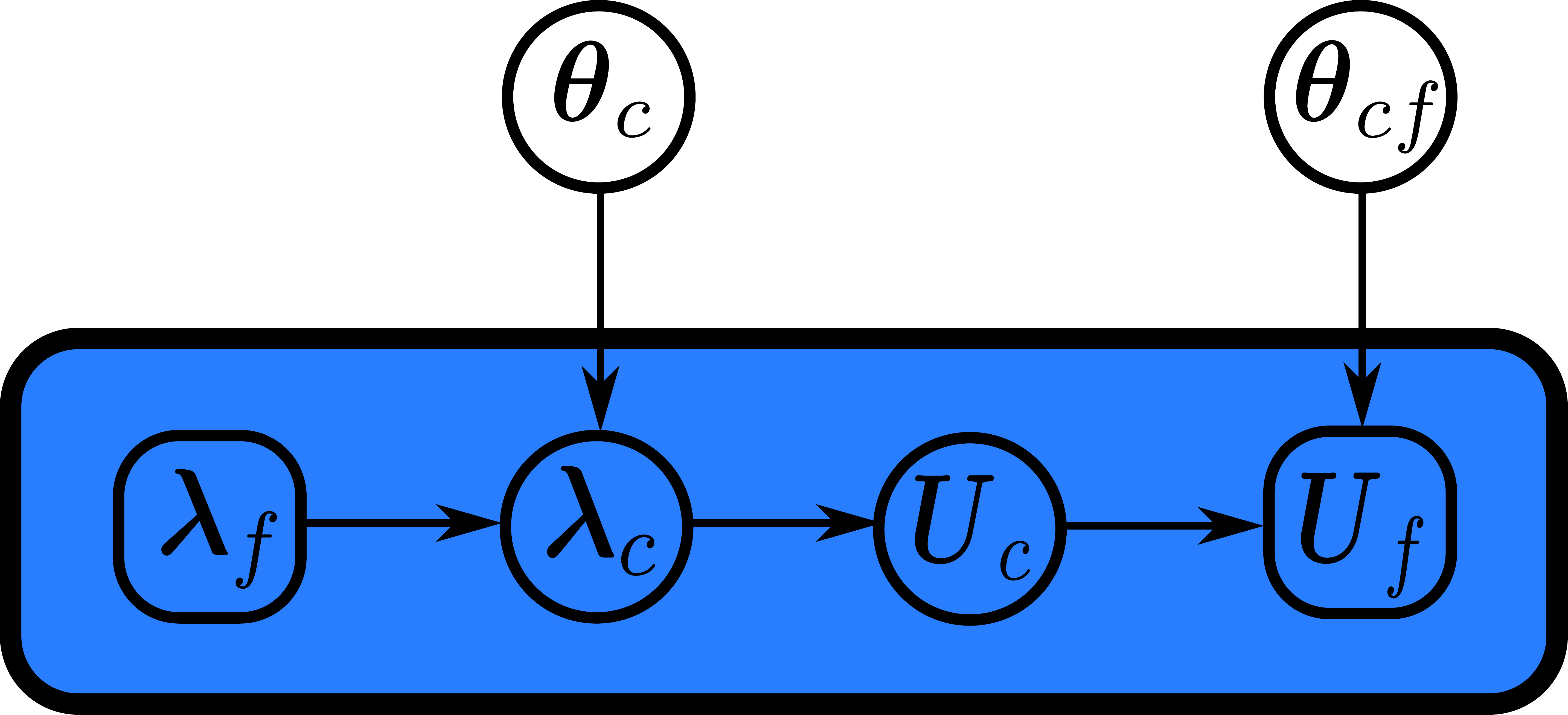}
\end{center}
\caption{A two-step Bayesian network/generative model defining $\bar{p}(\bs U_f|\bs 
\lambda_f, \bs \theta_c, \bs
\theta_{cf})$.}
\label{bnet}
\end{figure}
This model is illustrated graphically in Figure \ref{bnet}. The density  $p_c$  encodes $\bs \lambda_f$ into $\bs
\lambda_c$ and the coarse-to-fine map $p_{cf}$ plays the role of a decoder, i.e. given the CG output  $\bs U_c$, it
predicts $\bs U_f$.

Using the abbreviated notation  $\bs \theta = [\bs \theta_c, \bs \theta_{cf}]$, from \eqref{eq:autoencoder} we obtain
\be
\bar{p}(\bs U_f|\bs \lambda_f, \bs \theta) = \int p_{cf}(\bs U_f|\bs 
U_c(\bs \lambda_c), \bs
\theta_{cf})p_c(\bs \lambda_c|\bs \lambda_f, \bs \theta_c)d\bs\lambda_c,
\label{pbar}
\ee
where $\bs U_c(\bs \lambda_c)$ is the solution vector to equation \eqref{CG}. The previous discussion suggests the 
following 
generative process for drawing samples from $\bar{p}(\bs U_f|\bs \lambda_f, \bs  \theta)$ i.e. predicting the FG output
$\bs U_f$ given a FG input $\bs \lambda_f$,
\bi
\item draw a sample $\bs \lambda_c \sim p_c(\bs \lambda_c|\bs \lambda_f, \bs \theta_c)$,
\item solve the CG to obtain  $\bs U_c(\bs \lambda_c)$,
\item draw a sample $\bs U_f \sim p_{cf}(\bs U_f|\bs U_c(\bs \lambda_c), \bs \theta_{cf})$.
\ei

\subsection{Model training}
In order to train the model described above, it is a reasonable strategy to minimize the Kullback-Leibler divergence
\cite{Cover2012} between the target density $p(\bs U_f, \bs \lambda_f ) = \delta(\bs U_f - \bs  U_f(\bs \lambda_f))$ 
and
$\bar{p}(\bs U_f|\bs \lambda_f, \bs \theta)$. As these are conditional distributions, the KL divergence would depend on
$\bs \lambda_f$. In order to calibrate the model for the $\bs \lambda_f$ values that are of significance, we operate on 
the
augmented densities $p(\bs U_f,\lambda_f) =p(\bs U_f|\bs \lambda_f)p(\bs \lambda_f)$ and $\bar{p}(\bs U_f, \bs 
\lambda_f | \bs \theta)=\bar{p}(\bs U_f|\bs 
\lambda_f, \bs \theta)p(\bs
\lambda_f)$, where $p(\bs \lambda_f)$ is defined by \eqref{eq:inputpdf}. In particular, we propose minimizing with
respect to $\bs \theta$
\be
\begin{split}
\tm{KL}\left(p(\bs U_f, \bs \lambda_f ) || \bar{p}(\bs U_f, \bs \lambda_f ) \right) 
 & = ~\tm{KL}\left(p(\bs U_f, \bs \lambda_f) || \bar{p}(\bs U_f| \bs \lambda_f, \bs \theta) p(\bs \lambda_f) \right)\\
& = \int p(\bs U_f, \bs \lambda_f)\log\left(\frac{p(\bs U_f, \bs \lambda_f)}{\bar{p}(\bs U_f| \bs \lambda_f, \bs
\theta)p(\bs \lambda_f)} \right)d\bs U_fd\bs\lambda_f\\
 & = -\int p(\bs U_f, \bs \lambda_f) \log \bar{p}(\bs U_f|\bs \lambda_f, \bs \theta) d\bs U_f d\bs \lambda_f + H(p(\bs
 U_f, \bs \lambda_f))\\
& \approx  -\frac{1}{N} \sum_{i = 1}^N \log\bar{p}(\bs U_f^{(i)}|\bs \lambda_f^{(i)}, \bs \theta) + H(p(\bs U_f, \bs
\lambda_f)),
\end{split}
\label{eq:kl}
\ee
where $N$ is the number of training samples drawn from $p(\bs U_f, \bs \lambda_f)$, i.e. 
\be
\bs \lambda_f^{(i)} \sim p(\bs \lambda_f), \qquad \bs U_f^{(i)} = \bs U_f(\bs \lambda_f^{(i)}),
\ee
and $H(p(\bs U_f, \bs \lambda_f))$ is the entropy of $p(\bs U_f, \bs \lambda_f))$ that is nevertheless independent of
the model parameters $\bs \theta$. It is obvious from the final expression in \eqref{eq:kl} that     $ \sum_{i = 1}^N 
\log\bar{p}(\bs U_f^{(i)}|\bs \lambda_f^{(i)}, \bs \theta_c, \bs \theta_{cf})$ is a $\log$-likelihood function of the
data $\mathcal{D}_N$ which we denote by $L(\mathcal{D}_N|\bs \theta_c, \bs \theta_{cf})$. In a fully Bayesian setting,
this can be complemented with a prior $p(\bs \theta_c, \bs \theta_{cf})$ leading to the posterior
\be
p(\bs \theta_c, \bs \theta_{cf}| \mathcal{D}_N) \propto e^{L(\mathcal{D}_N|\bs \theta_c, \bs \theta_{cf})} p(\bs
\theta_c, \bs \theta_{cf}).
\label{posterior}
\ee
It is up to the analyst if predictions using equation \eqref{pbar} are carried out using point estimates of $\bs \theta$
(e.g. maximum likelihood (MLE) or maximum a posteriori (MAP)) or if a fully Bayesian approach is followed by  averaging
over the posterior  $p(\bs \theta_c, \bs \theta_{cf}| \mathcal{D}_N)$. The latter has the added advantage of 
quantifying the uncertainty introduced due the finite training data $N$. We pursue the former in the
following as it is computationally more efficient.

\subsubsection{Maximizing the posterior}
Our objective is to find $\bs \theta^*=[\bs \theta_c^*, \bs \theta_{cf}^*]$ which maximizes the  posterior given in
equation \eqref{posterior}, i.e.
\be
\begin{split}
[\bs \theta_c^*, \bs \theta_{cf}^*] &= \arg\max_{\bs \theta_c, \bs \theta_{cf}} 
e^{L(\mathcal{D}_N|\bs \theta_c, \bs \theta_{cf})} ~p(\bs
\theta_c, \bs \theta_{cf})\\
&= \arg\max_{\bs \theta_c, \bs \theta_{cf}}\left( L(\mathcal{D}_N|\bs \theta_c, \bs 
\theta_{cf}) + \log p(\bs \theta_c, \bs
\theta_{cf})\right)\\
&= \arg\max_{\bs \theta_c, \bs \theta_{cf}}\left( \sum_{i = 1}^N \log\bar{p}(\bs 
U_f^{(i)}|\bs \lambda_f^{(i)},
\bs \theta_c, \bs \theta_{cf}) + \log p(\bs \theta_c, \bs \theta_{cf})\right).
\end{split}
\label{MAP}
\ee
The main difficulty  in this optimization problem arises from the log-likelihood term  which involves the 
log of an analytically
intractable integral with respect to $\bs \lambda_c$ since
\be
\begin{split}
L_i(\bs \theta_c, \bs \theta_{cf}) &= \log\bar{p}(\bs U_f^{(i)}|\bs \lambda_f^{(i)}, \bs \theta_c, \bs \theta_{cf})\\
&= \log \int  p_{cf}(\bs U_f^{(i)}|\bs U_c(\bs \lambda_c^{(i)}), \bs \theta_{cf})p_c(\bs \lambda_c^{(i)}|\bs
\lambda_f^{(i)}, \bs \theta_c) ~d\bs\lambda_c^{(i)},
\end{split}
\ee
where we note that an independent copy of $\bs \lambda_c^{(i)}$ pertains to each data point $i$. Due to this
integration, typical deterministic optimization algorithms are not applicable.

However, as $\bs \lambda_c$ appears as a latent variable, we may resort to the well-known Expectation-Maximization
algorithm \cite{Dempster1977}. 
Using Jensen's inequality, we establish a lower bound on every single term $L_i$ of the sum in the log-likelihood
$L(\mathcal{D}_N|\bs \theta_c, \bs \theta_{cf})$ by employing an arbitrary density $ q_i(\bs \lambda_c^{(i)})$ 
(different for each sample $i$) as
\be
\begin{split}
L_i(\bs \theta_c, \bs \theta_{cf}) &= \log\bar{p}(\bs U_f^{(i)}|\bs \lambda_f^{(i)}, \bs \theta_c, \bs \theta_{cf})\\
&=\log \int p_{cf}(\bs U_f^{(i)}|\bs U_c(\bs \lambda_c^{(i)}), \bs \theta_{cf})p_c(\bs \lambda_c^{(i)}|\bs
\lambda_f^{(i)}, \bs \theta_c)d\bs\lambda_c^{(i)}\\
&=\log \int q_i(\bs \lambda_c^{(i)}) \frac{p_{cf}(\bs U_f^{(i)}|\bs U_c(\bs \lambda_c^{(i)}), \bs \theta_{cf})p_c(\bs
\lambda_c^{(i)}|\bs \lambda_f^{(i)}, \bs \theta_c)}{q_i(\bs \lambda_c^{(i)})}d\bs\lambda_c^{(i)}\\
&\ge \int q_i(\bs \lambda_c^{(i)}) \log\left(\frac{p_{cf}(\bs U_f^{(i)}|\bs U_c(\bs \lambda_c^{(i)}), \bs
\theta_{cf})p_c(\bs \lambda_c^{(i)}|\bs \lambda_f^{(i)}, \bs \theta_c)}{q_i(\bs
\lambda_c^{(i)})}\right)d\bs\lambda_c^{(i)} ~ \textrm{(Jensen)}\\
&=\mathcal F_i(q_i; \bs \theta_c, \bs \theta_{cf}),
\end{split}
\label{eq:lb}
\ee
Hence, the $\log$-posterior in \eqref{posterior} can be lower bounded by
\be
\begin{split}
\log p(\bs \theta_c, \bs \theta_{cf}| \mathcal{D}_N)& = \log  L(\mathcal{D}_N|\bs 
\theta_c, \bs \theta_{cf}) + \log p(\bs \theta_c,\bs \theta_{cf})\\
&= \sum_{i = 1}^N \log L_i(\bs \theta_c, \bs \theta_{cf}) + \log p(\bs \theta_c, \bs \theta_{cf})\\
&\ge \sum_{i = 1}^N \mathcal F_i(q_i;\bs \theta_c, \bs \theta_{cf}) + \log p(\bs \theta_c, \bs \theta_{cf})\\
&= \mathcal F\left(\left\{q_i\right\}_{i = 1}^N, \bs \theta_c, \bs \theta_{cf}\right) + \log p(\bs \theta_c, \bs
\theta_{cf}).
\end{split}
\label{Jensen}
\ee
The introduction of the auxiliary densities $q_i$ suggests the following  recursive procedure
\cite{Bishop2006} for maximizing the log-posterior:
\bi
\item[\tbf{E-step:}] Given some $\bs \theta_c^{(t)}, \bs \theta_{cf}^{(t)}$ in iteration $t$, find the auxiliary
densities $q_i^{(t + 1)}$ that maximize $\mathcal F$,
\item[\tbf{M-step:}] Given $q_i^{(t + 1)}$, find the parameters $\bs \theta_c^{(t + 1)}, \bs \theta_{cf}^{(t + 1)}$ that
maximize $\mathcal F$.
\ei

It can be readily shown that the optimal $q_i$ is given by
\be
q_i(\bs \lambda_c^{(i)}) \propto p_{cf}(\bs U_f^{(i)}|\bs U_c(\bs \lambda_c^{(i)}), \bs \theta_{cf})p_c(\bs
\lambda_c^{(i)}|\bs \lambda_f^{(i)}, \bs \theta_c)
\label{qopt}
\ee
with which the inequality in \eqref{eq:lb} becomes an equality. In fact, both E- and M-steps can be relaxed to find 
suboptimal
$q^{(t)}_i, \bs \theta_c^{(t)}, \bs \theta_{cf}^{(t)}$, which enables the application of approximate schemes such
as e.g. Variational Inference (VI) \cite{Wainwright2008, Hoffman2013}.
For the M-step,  we may resort to any (stochastic) optimization
algorithm \cite{robbins_stochastic_1951} or, on occasion, closed-form updates might also be feasible. 

\newpage
\section{SAMPLE PROBLEM: 2D STATIONARY HEAT EQUATION}
\begin{figure}[t]
\begin{center}
\includegraphics[width=\textwidth]{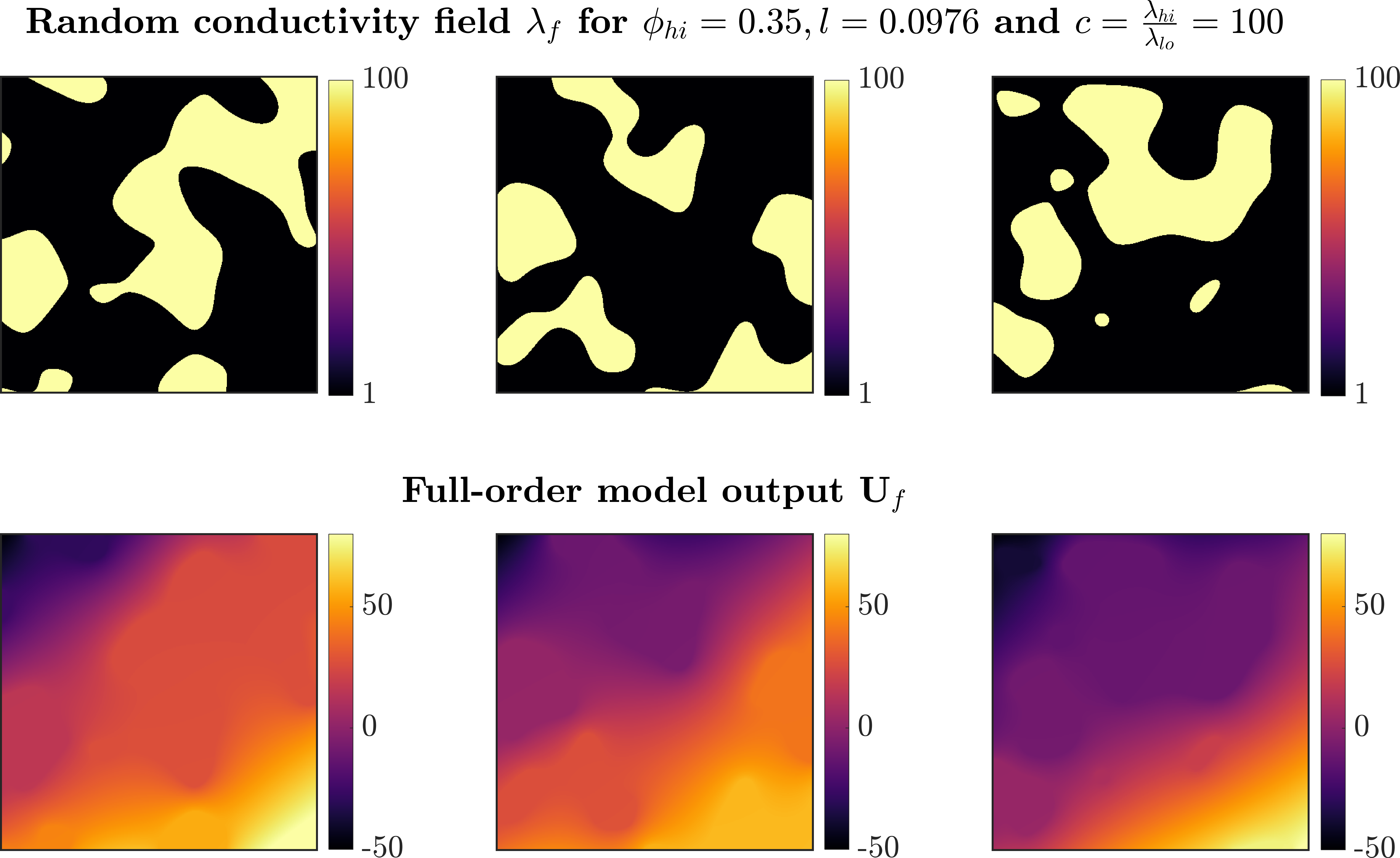}
\end{center}
\caption{Random microstructure samples and corresponding fine-grained model outputs.}
\label{FGsamples}
\end{figure}
As a sample problem, we consider the 2D stationary heat equation on the unit square $[0,1]^2$
\be
\nabla_{\bs x}(-\lambda(\bs x, \xi(\bs x))\nabla_{\bs x}U(\bs x, \xi(\bs x))) = 0 
\label{heatequation}
\ee
where $U(\bs x, \xi(\bs x))$ represents the temperature field. For the boundary conditions, we fix the temperature $U$
in the upper left corner (see Figure \ref{FGsamples}) to $-50$ and prescribe the heat flux $\bs Q(\bs x) = \bsmat 150
-30y
\\
100 -30x \esmat$ on the remaining domain boundary $\pa \mathcal D$.

We consider a binary random medium whose conductivity $\lambda(\bs x)$ can take the values $\lambda_{\tm{hi}}$ and
$\lambda_{\tm{lo}}$. To define such a field we consider transformations of a zero-mean Gaussian process $\xi(\bs x)$
of the form \cite{Roberts1995, Koutsourelakis2005}
\be
\lambda(\bs x, \xi(\bs x)) = \begin{cases}
 \lambda_{\tm{hi}},  \qquad \tm{if}\qquad \xi(\bs x) > c, \\
\lambda_{\tm{lo}}, \qquad \tm{otherwise}
\end{cases}
\label{lambdaRand}
\ee
where the thresholding constant  $c$ is selected so as to achieve the target volume fraction  $\phi_{\tm{hi}}$ (or
equivalently $\phi_{\tm{lo}}=1-\phi_{\tm{hi}}$) of the material with conductivity $\lambda_{\tm{hi}}$ (or equivalently
of $\lambda_{\tm{lo}}$). In the following we use an isotropic squared-exponential covariance function for $\xi(\bs x)$
of the form
\be
\tm{cov}(\bs x_i, \bs x_j) = \exp\left\{-\frac{|\bs x_i - \bs x_j|^2}{l^2} \right\}.
\label{GP}
\ee
In the following studies,  we used values $l \approx 0.01$. 

Due to the small correlation length $l$, we discretize the SPDE  in Equation \eqref{heatequation} using $256\times
256$ standard quadrilateral finite elements, leading to a linear system of $N_{\tm{eq}} = \dim(\bs U_f) - 1 = 257\times 
257 - 1 = 66048$ algebraic
equations. We choose the discretization mesh of the random process $\lambda(\bs x, \xi(\bs x))$ to coincide with the
finite element discretization mesh of the SPDE, i.e. $\dim(\bs \lambda_f) = 256\times 256 = 65536$.

Samples $\mathcal{D}_N=\{\lambda_f^{(i)}, \bs U_f^{(i)}\}_{i=1}^N$ are readily obtained by simulating realizations of
the discretized Gaussian field, transforming them according to \eqref{lambdaRand} and solving the discretized
SPDE. Three such samples can be seen in Figure  \ref{FGsamples}.

\subsection{Model specifications}

We define a coarse model employing $n_{\bs \lambda_c}$ quadrilateral elements, the 
conductivities of which are given by the vector $\bs \lambda_c$.  Since these need to be strictly positive, we operate 
instead on $\bs{z}_c$ defined as
\be
\bs z_c = \log \bs \lambda_c.
\ee
For each element $k=1,\ldots, n_{\bs \lambda_c}$ of the coarse model/mesh, we assume that
\be
z_{c, k} = \sum_{j = 1}^{N_{\tm{features}}} \theta_{c, j}\chi_j(\bs \lambda_{f, k}) + \sigma_k Z_k, \qquad Z_k \sim
\mathcal N(0, 1),
\label{linModPc}
\ee
where $\bs \lambda_{f, k}$ is the subset of the vector $\bs \lambda_f$ that belongs to coarse element $k$ and 
$\chi_j$
some feature functions of the underlying $\bs \lambda_{f, k}$ that we  specify below. In a more compact form, we can 
write
\be
p_c(\bs z_c| \bs \lambda_f, \bs \theta_c, \bs \sigma) = \mathcal N(\bs z_c| \bs \Phi(\bs \lambda_f)\bs \theta_c,
\tm{diag}(\bs \sigma^2)),
\label{p_c}
\ee
where $\bs \Phi(\bs \lambda_f)$ is an $n_{\bs \lambda_c} \times N_{\tm{features}}$ design matrix with $ \Phi_{kj}(\bs
\lambda_f) = \chi_j(\bs \lambda_{f, k})$ and $\tm{diag}(\sigma^2)$ is a diagonal covariance matrix with components
$\sigma_k^2$.

For the coarse-to-fine map $p_{cf}$, we postulate the relation
\be
p_{cf}(\bs U_f|\bs U_c(\bs z_c), \bs \theta_{cf}) = \mathcal N(\bs U_f|  \bs W \bs U_c(\bs z_c), \bs S),
\label{p_cf}
\ee
where $\bs \theta_{cf} = ( \bs W, \bs S)$ are parameters to be learned from the data. The matrix $\bs{W}$ is of
dimension $n_f \times n_c$ and $\bs S$ is a positive definite covariance matrix of size $n_f \times n_f$.
To reduce the  large amount of free parameters in the model, we fix $\bs{W}$ to express coarse model's shape functions.
Furthermore, we assume that $\bs{S} = \tm{diag}(\bs s)$ where $\bs s$ is the $n_f$-dimensional vector of diagonal
entries of the diagonal matrix $\bs S$. The aforementioned expression implies, on average, a linear relation between the
fine and coarse model outputs, which is supplemented by the residual Gaussian noise implied by $\bs{S}$. The latter
expresses the uncertainty in predicting the FG output when only the CG solution is available.   We note that the 
relation in \refeq{linModPc} (i.e. $\btc$ and $\sigma_k^2$) will be adjusted during training so that the model implied 
in \refeq{p_cf} represents the data as good as  possible.

From equation \eqref{qopt}, we have that
\be
\begin{split}
\log q_i^{(t + 1)}(\bs z_c^{(i)}) &\propto \frac{1}{2}\sum_{k = 1}^{N_{\tm{el}}^c}\log
\left((\sigma_k^{(t)})^{-2}\right) -\frac{1}{2}\sum_{k = 1}^{N_{\tm{el}}^c}(\sigma_k^{(t)})^{-2} \left(\bs z_c^{(i)} -
\bs \Phi(\bs \lambda_f^{(i)})\bs \theta_c^{(t)}\right)_k^2\\
&- \frac{1}{2}\sum_{j = 1}^{N_{\tm{el}}^f}\log s_j^{(t)} - \frac{1}{2}\sum_{j =
1}^{N_{\tm{el}}^f}s_j^{(t)}\left(\bs T_f^{(i)} - \bs W^{(t)}\bs T_c(\bs z_c^{(i)}) \right)_j^2,
\end{split}
\ee
where with $(.)_i$, we mean the $i$-th component of the vector in brackets. For use in the M-step, we compute the
gradients
\begin{align}
\frac{\pa \mathcal F}{\pa \sigma^{-2}_k} &= \frac{N}{2}\sigma^2_k - \frac{1}{2}\sum_{i = 1}^N\left< \left(\bs z_c^{(i)}
- \bs \Phi(\bs \lambda_f^{(i)})\bs \theta_c^{(t)}\right)_k^2\right>_{q_i},\\
\nabla_{\bs \theta_c}\mathcal F &= \sum_{i = 1}^N \left(\bs \Phi^T(\bs z_f^{(i)}) \bs \Sigma^{-1}\left<
\bs z_c^{(i}\right>_{q_i^{(t)}} - \bs \Phi^T(\bs \lambda_f^{(i)})\bs \Sigma^{-1}\bs\Phi(\bs \lambda_f^{(i)}) \bs
\theta_c\right)
\end{align}
where $\bs \Sigma = \tm{diag}(\sigma_1^2,\ldots,\sigma^2_{n_{\bs \lambda_c}})$. We can readily compute the roots to find
the optimal $\bs \Sigma, \bs \theta_c$. Furthermore, from$\frac{\pa \mathcal F}{\pa s_j} = 0$ we get
\be
s_j^{(t + 1)} = \frac{1}{N}\sum_{i = 1}^N \left<\left(\bs T_f^{(i)} - \bs W\bs 
T_c(\bs z_c^{(i)}) \right)^2_j
\right>_{q_i^{(t)}}.
\ee

\subsection{Feature functions \texorpdfstring{$\chi$}{}}
The framework advocated allows for any form and any number of feature functions $\chi_j$ in \eqref{linModPc}.
Naturally such a selection can be guided by physical insight \cite{Koutsourelakis2006}. In practice, the feature
functions we are using can be roughly classified into two different groups:

\subsubsection{Effective-medium approximations}
We consider existing effective-medium approximation formula that can be found in the literature \cite{Torquato2001} as
ingredients for construction of feature functions $\chi_j$ in equation \eqref{linModPc}. The majority of commonly used
such features only retain low-order topological information. 
only 
The following approximations to the effective property
$\lambda_{\tm{eff}}$ can be used as building blocks for feature functions $\chi$ in the sense that we can transform
them nonlinearly such that $\chi(\bs \lambda_f) = f(\lambda_{\tm{eff}}(\bs \lambda_f))$. In particular, as we are
modeling $\bs z_c = \log \bs \lambda_c$, we include feature functions of type $\chi(\bs \lambda_f) =
\log(\lambda_{\tm{eff}}(\bs \lambda_f))$.


\paragraph{Maxwell-Garnett approximation (MGA)}
The Maxwell-Garnett approximation is assumed to be valid if the microstructure consists of a matrix phase
$\lambda_{\tm{mat}}$ and spherical inclusions $\lambda_{\tm{inc}}$ that are small and dilute enough such that 
interactions between them can be
neglected. 
another
Moreover, the heat flux far away from any inclusion is assumed to be constant. Under such 
conditions, the effective
conductivity can be approximated in 2D as
\be
\lambda_{\tm{eff}} = \lambda_{\tm{mat}}\frac{\lambda_{\tm{mat}} + \lambda_{\tm{inc}} + 
\phi_{\tm{inc}}(\lambda_{\tm{inc}} - 
\lambda_{\tm{mat}})}{\lambda_{\tm{mat}} +
\lambda_{\tm{inc}} - \phi_{\tm{inc}}(\lambda_{\tm{inc}} - \lambda_{\tm{mat}})},
\ee
where the volume fraction $\phi_i(\bs \lambda_f)$ is given by the fraction of phase $i$ elements in the binary vector
$\bs \lambda_f$.

\paragraph{Self-consistent approximation (SCA)}
The self-consistent approximation (or Bruggeman \\ formula) was originally developed for effective electrical properties
of random microstructures. It also considers non-interacting spherical inclusions and follows from the assumption that
perturbations of the electric field due to the inclusions average to $0$. In 2D, the formula reads
\be
\lambda_{\tm{eff}} = \frac{\alpha + \sqrt{\alpha^2 + 4\lambda_{\tm{mat}} \lambda_{\tm{inc}}}}{2}, \qquad 
\alpha = \lambda_{\tm{mat}}(2\phi_{\tm{mat}} - 1) +
\lambda_{\tm{inc}}(2\phi_{\tm{inc}} - 1).
\ee
Note that the SCA exhibits phase inversion symmetry.

\paragraph{Differential effective medium (DEM)}
From a first-order expansion in volume fraction of the effective conductivity in the dilute limit of spherical
inclusions, one can deduce the differential equation \cite{Torquato2001}
\be
(1 - \phi_{\tm{inc}})\frac{d}{d\phi_{\tm{inc}}}\lambda_{\tm{eff}}(\phi_{\tm{inc}}) = 
2\lambda_{\tm{eff}}(\phi_{\tm{inc}})\frac{\lambda_{\tm{inc}} -
\lambda_{\tm{eff}}(\phi_{\tm{inc}})}{\lambda_{\tm{inc}} + \lambda_{\tm{eff}}(\phi_{\tm{inc}})},
\ee
which can be integrated to
\be
\left(\frac{\lambda_{\tm{inc}} - \lambda_{\tm{eff}}}{\lambda_{\tm{inc}} - \lambda_{\tm{mat}}} \right) 
\sqrt{\frac{\lambda_{\tm{mat}}}{\lambda_{\tm{eff}}}} =
1 - \phi_{\tm{inc}}.
\ee
We solve for $\lambda_{\tm{eff}}$ and use it as a feature function $\chi$.

\subsubsection{Morphology-describing features}
Apart from the effective-medium approximations which only take into account the phase conductivities and 
volume fractions, we
wish to have feature functions $\chi_j$ that more thoroughly describe the morphology of the underlying
microstructure. Popular members of this class of microstructural features are the two-point correlation, the
lineal-path function, the pore-size density or the specific surface to mention only a few of them \cite{Torquato2001}.

We are however free to use any function $\chi_j: (\mathbb{R}^+)^{\dim(\bs \lambda_{f, k})} \mapsto \mathbb{R}$ as a
feature, no matter from which field or consideration it may originate. We thus make use of existing image processing
features \cite{MATLAB:2016} as well as novel topology-describing functions. Some important examples are

\begin{figure}[t]
\begin{center}
\includegraphics[width=\textwidth]{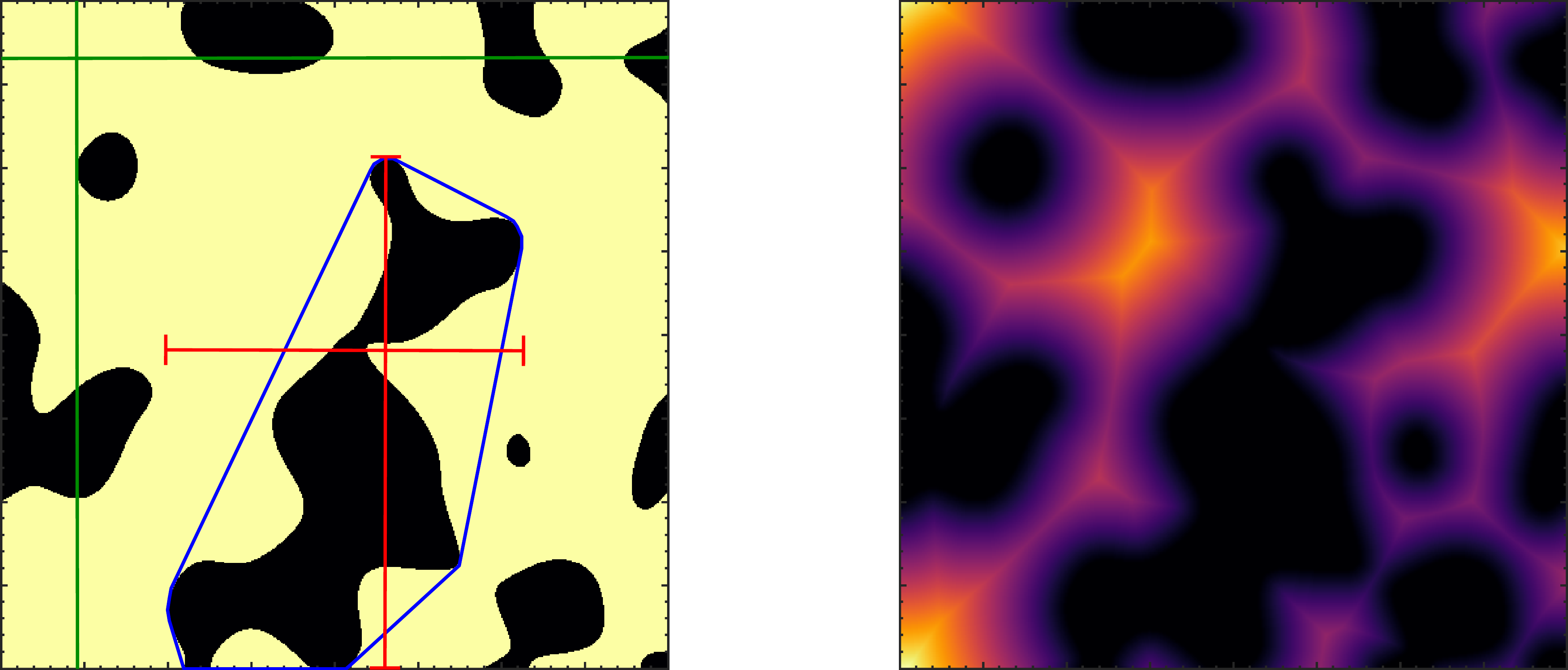}
\end{center}
\caption{Left: Sample microstructure for $l = 0.078$ and $\phi_{\tm{hi}} = 0.7$. The blue line encompasses the convex
area of the encircled low conducting phase blob, the two red lines are its maximum extent in $x$- and $y$-direction.
The green lines are paths along which we count pixels and compute generalized means in the pixel-cross and straight path
mean functions. Right: distance transform (distance to nearest black pixel) of the microstructure.}
\label{sampleMicrostructure}
\end{figure}

\paragraph{Convex area of connected phase blobs}
This feature identifies distinct, connected ``blobs'' of only high or low conducting phase pixels and computes the area
of the convex hull to each blob. We then can e.g. use the mean or maximum value thereof as a feature function.

\paragraph{Blob extent}
From an identified phase blob, we can compute its maximum extension in $x$- and $y$-directions. One can for instance
take the mean or maximum of maximum extension among identified blobs as a feature.

\paragraph{Distance transformation functions}
The distance transform of a binary image assigns a number to each pixel $i$ that is the distance from that pixel $i$ to
the nearest nonzero pixel in the binary image, see right part of Figure \ref{sampleMicrostructure}. One can use either
phase to correspond to nonzero in the binary image as well as different distance metrics. As a feature, one can e.g.
take the mean or maximum of the distance transformed image.

\paragraph{Pixel-cross function}
This feature counts the number of high or low conducting pixels one has to cross going on a straight line from boundary
to boundary in $x$- or $y$-direction. One can again take e.g. mean, maximum or minimum values as the feature function
outputs.

\paragraph{Straight path mean function}
A further refinement of the latter function is to take generalized means instead of numbers of crossed pixels along
straight lines from boundary to boundary. In particluar, we use harmonic, geometric and arithmetic means as features.

\subsection{Sparsity priors}
\label{sec:sparsityPrior}
It is clear from the previous discussion that the number of feature functions is practically limitless. The more such
$\chi_j$ one introduces, the more unknown parameters $\theta_{c,j}$ must be learned. From the modeling point of
view, while ML estimates can always be found, it is desirable to have as clear of a distinction as possible between
relevant and irrelevant features that could provide further insight as well being able to do so with the fewest
possible training data available. For that purpose we advocate the use of sparsity-enforcing priors in $\bs \theta_c$ 
\cite{Bernardo2003, Figueiredo2003a,DBLP:journals/jcphy/SchoberlZK17}. From a statistical perspective, this is also 
motivated  by the 
\tit{bias-variance-tradeoff}. Model prediction accuracy is adversely affected by two factors: One is noise in the 
training
data (\tit{variance}), the other is due to overly simple model assumptions (\tit{bias}). Maximum-likelihood estimates
of model parameters tend to have low bias but high variance, i.e. they accurately predict the training data but
generalize poorly. To address this issue, a common Bayesian approach to control model complexity is the use of priors,
which is the equivalent to regularization in frequentist formulations. A particularly appealing family of prior
distributions is the Laplacian (or LASSO regression, \cite{Tibshirani1996}), as it sets redundant or unimportant
predictors to exactly 0, thereby simplifying interpretation.

In particular, we use a prior on the coefficients $\bs \theta_c$ of the form
\be
\log p(\bs \theta_c) = \log \frac{\sqrt{\gamma}}{2} -\sqrt{\gamma}\sum_{i = 1}^{N_{\bs 
\theta_c}} \left|\theta_{c,i}
\right|,
\label{Laplace}
\ee
with a hyper-parameter $\gamma$ which can be set by either applying some cross-validation scheme or more efficiently by
minimization of Stein's unbiased risk estimate \cite{Stein1981, Zou2007}. A  straightforward implementation
of this prior is described in \cite{Figueiredo2003a}.

\section{RESULTS}
\subsection{Required training data}
\label{errorData}
We use a total number of 306 feature functions $\chi_j$ and a Laplacian prior \eqref{Laplace} with a hyperparameter
$\gamma$ we find by cross-validation. We set the length scale parameter to $l = 0.0781$ and the expected volume fraction
of the high conducting phase to $\phi_{\tm{hi}} = 0.2$. For the contrast, we take $c =
\frac{\lambda_{\tm{hi}}}{\lambda_{\tm{lo}}} = 10$, where we set $\lambda_{\tm{lo}} = 1$. The full- and reduced-order
models are both computed on regular quadrilateral finite element meshes of size $256\times 256$ for the FG and
$2\times2$, $4\times 4$ and $8\times 8$ for the CG.
\begin{figure}[t]
\centering
\includegraphics[width=\textwidth]{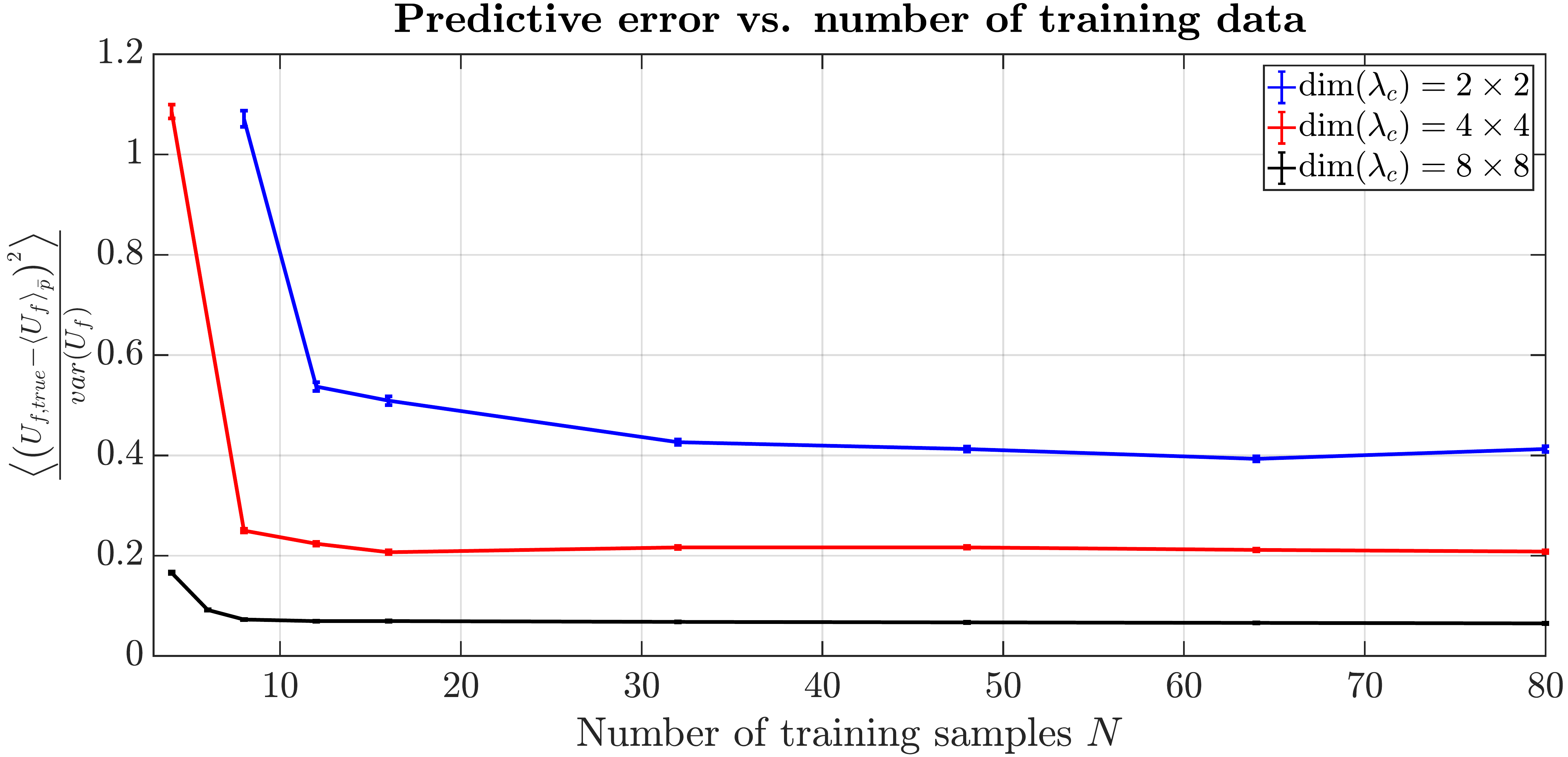}
\caption{Relative squared prediction error $\frac{\left<\left(\bs U_{f, \tm{true}} - \left<\bs U_f \right>_{\bar{p}}
\right)^2 \right>}{var(\bs U_f)}$ versus number training data samples $N$ and different CG mesh sizes $N_{\tm{el},
c} = \dim(\bs \lambda_c)$. We set $\phi_{\tm{hi}} = 0.2$, $l = 0.0781$ and $c = 10$.}
\label{errorPlot}
\end{figure}

Our goal is to measure the predictive capabilities of the described model. To that end, we compute the mean squared
distance
\be 
d^2 = \frac{1}{N_{\tm{test}} n_f}\sum_{j = 1}^{n_f}\sum_{i = 1}^{N_{\tm{test}}} \left(U_{f, \tm{true}, j}^{(i)} -
\left<U_{f, j}^{(i)} \right>_{\bar{p}(\bs U_f^{(i)})|\bs \lambda_f^{(i)}, \bs \theta)} \right)^2
\ee
of the predictive mean $\left<\bs U_{f}^{(i)} \right>_{\bar{p}(\bs U_f^{(i)})|\bs \lambda_f^{(i)}, \bs \theta)}$ to the
true FG output $\bs U_f^{(i)}$ on a test set of $N_{\tm{test}} = 256$ samples. Predictions are carried out by drawing
10,000 samples of $\bs \lambda_c$ from the learned distribution $p_c(\log \bs \lambda_c| \bs \lambda_f, \bs \theta_c^*,
\bs \sigma) = \mathcal N(\log(\bs \lambda_c)| \bs \Phi(\bs \lambda_f)\bs \theta_c^*, \tm{diag}(\bs \sigma^2))$, solving
the coarse model $\bs U_c^{(i)} = \bs U_c(\bs \lambda_c^{(i)})$ and drawing a sample from $p_{cf}(\bs U_f^{(i)}|\bs
U_c^{(i)}, \bs \theta_{cf}^*) = \mathcal N(\bs U_f^{(i)}|  \bs W \bs U_c^{(i)}, \bs S^*)$ for every test data point.
Monte Carlo noise is small enough to be neglected.

As a reference value for the computed error $d^2$, we compute the mean variance of the FG output
\be
var(U_f) = \frac{1}{n_f}\sum_{i = 1}^{n_f}\left(\left<U_{f,i}^{2} \right> - \left<U_{f,i} \right>^2\right),
\ee
where expectation values $\left<.\right>$ are estimated on a set of 1024 FG samples such that errors due to Monte Carlo
can be neglected. Figure \ref{errorPlot} shows the relative squared prediction error $\frac{d^2}{var(U_f)}$ in
dependence of the number of training data samples for different coarse mesh sizes $\dim(\bs \lambda_c) = 2\times 2,
4\times 4$ and $8\times 8$. We observe that the predictive error converges to a finite value already for relatively few
training data. This is due to the inevitable information loss during the coarsening process $\bs \lambda_f \rightarrow
\bs \lambda_c$ as well as finite element discretization errors. In accordance with that, we see that the error drops
with the dimension of the coarse mesh, $\dim(\bs \lambda_c)$.

\subsection{Activated microstructural feature functions}

\begin{figure}[t]
\begin{center}
\includegraphics[width=\textwidth]{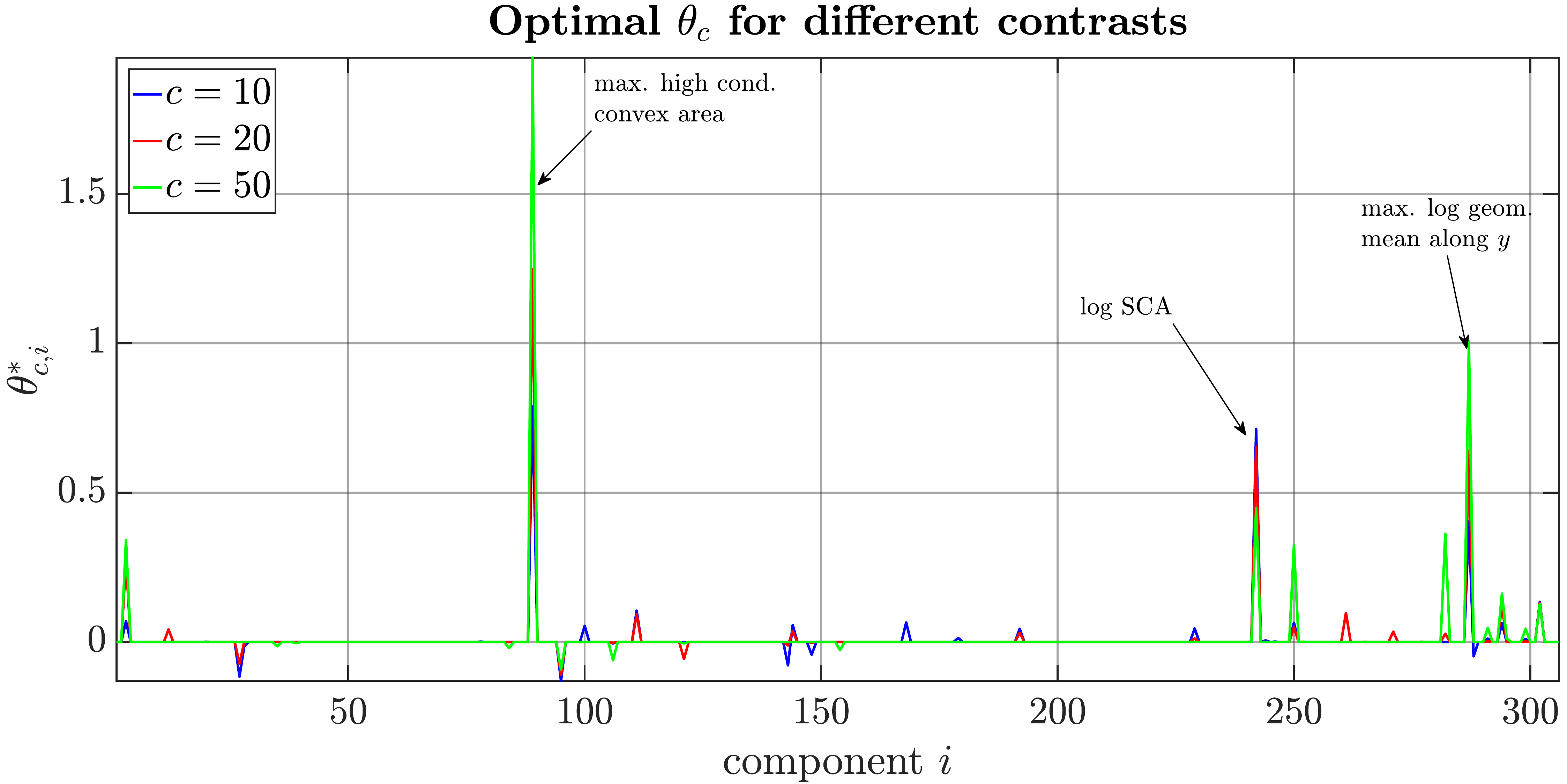}
\caption{Components of the optimal $\bs \theta_c^*$ for different values of contrast $c = \left\{10, 20, 50 \right\}$.
Most feature functions are deactivated by the sparsity prior. We observe that with increasing contrast, the importance
of the $\log$ SCA diminishes at the expense of more geometric features.}
\label{thetaOpt}
\end{center}
\end{figure}

For the same volume fraction and microstructural length scale parameter as above \\($\phi_{\tm{hi}} = 0.2, l = 0.0781$),
a varying contrast of $c = \left\{10, 20, 50 \right\}$, a coarse model dimension of $\dim(\bs \lambda_c) = 4\times 4$ and
a training set of $N = 128$ samples, we find the optimal $\bs \theta_c^*$ shown in Figure \ref{thetaOpt}.
Due to the application of a sparsity enforcing prior as described in section \ref{sec:sparsityPrior}, we observe that
most components of $\bs \theta_c^*$ are exactly 0. Comparability between different feature functions can be ensured by
standardization or normalization of feature function outputs on the training data. For all three contrast values, we see
that the three most important feature functions are given by the maximum convex area of  blobs of conductivity
$\lambda_{\tm{hi}}$, the maximum $\log$ of the geometric mean of conductivities along a straight line from boundary to
boundary in $y$-direction and the $\log$ of the self-consistent effective medium approximation as described above.
The  maximum convex area feature returns   the largest convex area of all blobs found within a coarse element.
convex area. 
It is an interesting to note  that although, in the set of 306 feature functions $\chi_j$,  the 
$(\max/\min/$mean/variance
of) the blob area (for both high and low conducting phases) are also included, it is only the  convex area which is 
activated.

In Figure \ref{thetaOpt}, it is observed that with increasing contrast, the coefficient $\theta_{c,j}^*$ belonging to
the $\log$ self-consistent approximation is decreasing in contrast to increasing values of 
$\theta_{c,j}^*$'s corresponding the $\max.$ high conducting convex area and the $\max.$ $\log$ geometric mean along
$y$-direction. We believe that this is because, the higher the contrast  $c$ is, the more the exact 
geometry and connectedness
of the microstructure plays a role for predicting the effective properties. As the SCA only considers the volume 
fraction
and the conductivities of both phases, it disregards such information.

\subsection{Learned effective property \texorpdfstring{$\bs \lambda_c$}{}}
\begin{figure}[t]
\centering
\includegraphics[width=\textwidth]{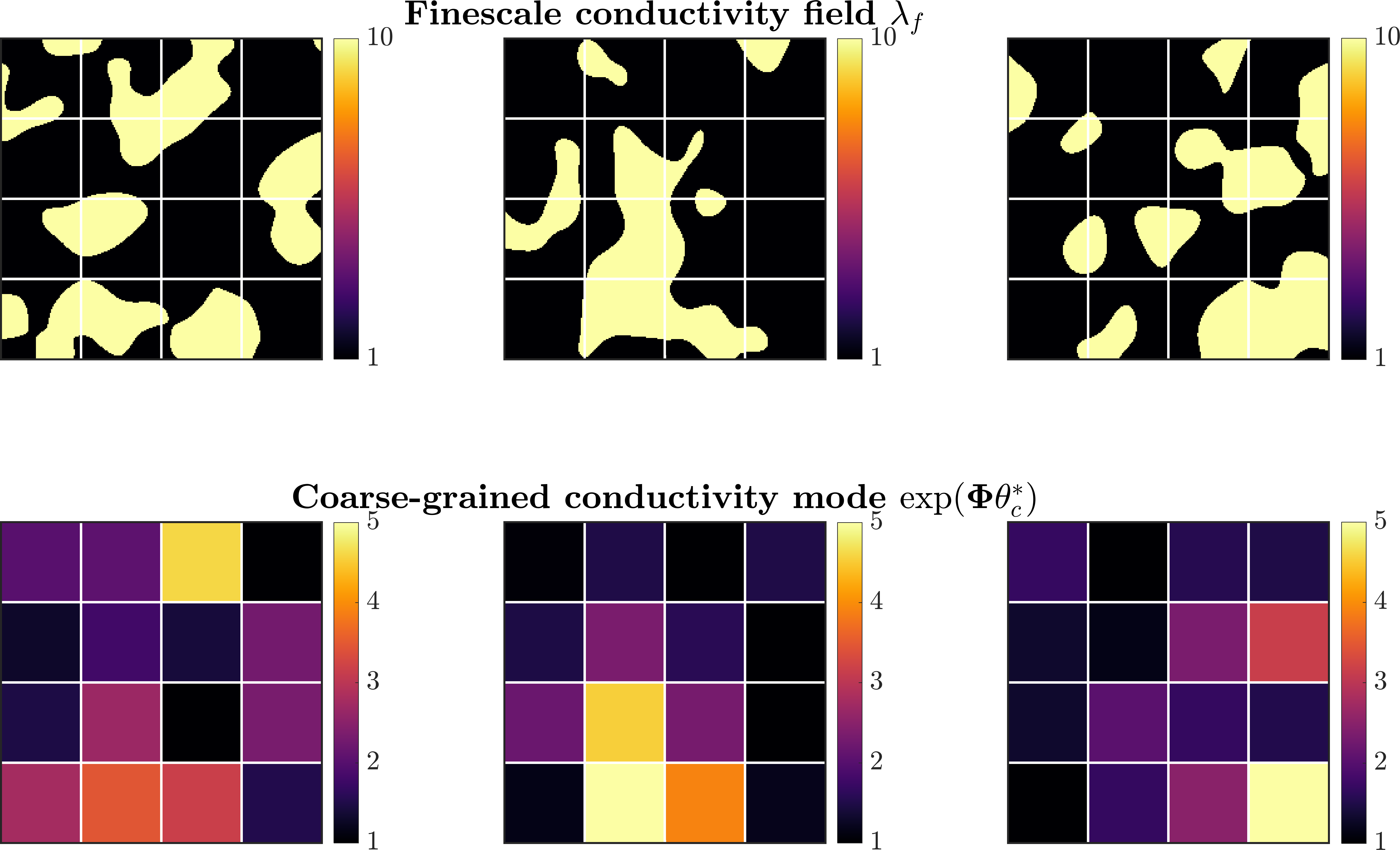}
\caption{Three test samples and the corresponding mode $\exp(\bs \Phi(\bs \lambda_f)\bs \theta_c^*)$ of $p_c$.}
\label{effCond}
\end{figure}
Figure \ref{effCond} shows three test samples of $\phi_{\tm{hi}} = 0.2, l = 0.0781$ and $c = 10$ (top row) along with
the mode $\exp\left(\bs \Phi(\bs \lambda_f)\bs \theta_c^*\right)$ of the learned distribution of the effective
conductivity $p_c(\bs \lambda_c|\bs \lambda_f, \bs \theta_c^*)$, which is connected to the mean by $\left<\bs \lambda_c
\right>_{p_c} = \exp\left(\bs \Phi(\bs \lambda_f)\bs \theta_c^* + \frac{1}{2}\bs \sigma^{*2}\right)$. We emphasize again
that the latent variable $\bs \lambda_c$ is not a lower-dimensional compression of $\bs \lambda_f$ with the objective
of most accurately reconstructing $\bs \lambda_f$, but of providing good predictions of $\bs U_f(\bs \lambda_f)$. Even
though Figure \ref{effCond} gives the impression of a simple local averaging relation between $\bs \lambda_f$ and $\bs
\lambda_c$, this is not always be the case. In particular, $p_c(\bs \lambda_c|\bs\lambda_f, \bs \theta_c^*)$ was found 
to have  non-vanishing probability mass for $\lambda_{c,i} < \lambda_{\tm{lo}}$ or $\lambda_{c,i} > 
\lambda_{\tm{hi}}$,
especially for more general models where the coarse-to-fine mapping \eqref{p_cf} is not fixed to be the shape function
interpolation $\bs W$.

\newpage
\subsection{Predictive uncertainty}
\begin{figure}[h]
\centering
\includegraphics[width=\textwidth]{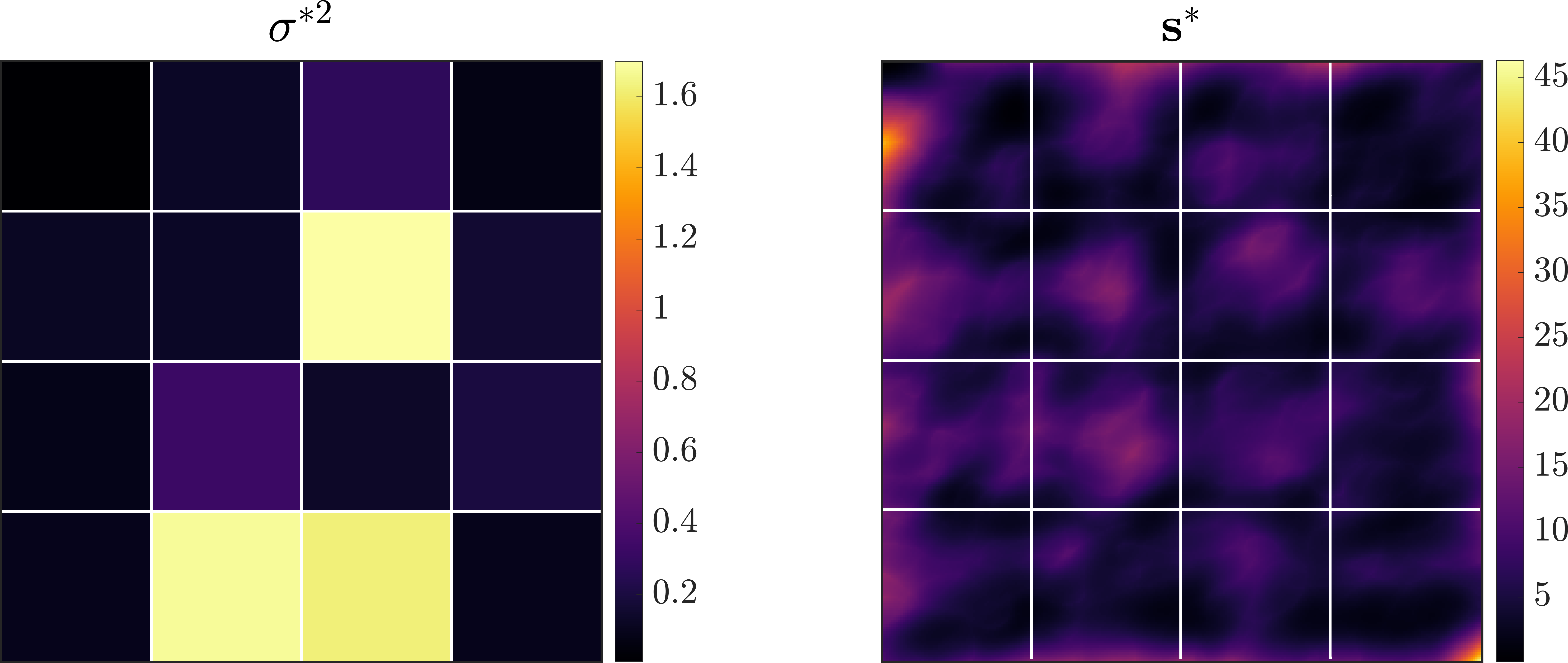}
\caption{Variance parameters $\bs \sigma^{*2}$ and $\bs s^*$ of $p_c$ and $p_{cf}$, respectively.}
\label{uncertainty}
\end{figure}
The predictive uncertainty is composed of the uncertainty in having an accurate encoding of $\bs \lambda_f$ in $\bs
\lambda_c$ which is described by $\sigma_k^2$ in $p_c$ \eqref{p_c}, as well as the uncertainty in the reconstruction
process from $\bs U_c$ to $\bs U_f$, which is given by the diagonal covariance $\bs S = \tm{diag}(\bs s)$ in $p_{cf}$
\eqref{p_cf}. Both are shown after training ($\phi_2 = 0.2, l = 0.0781, c = 10$) in Figure \ref{uncertainty}. For
$\bs\sigma^{*2}$, we observe that values in the corner elements always converge to very tight values, whereas some
non-corner elements can converge to comparably large values. The exact location of these elements is data dependent.
The coarse-to-fine reconstruction variances $\bs s^*$ is depicted on the right column  of Figure \ref{uncertainty}. As 
expected,
we see that the estimated coarse-to-fine reconstruction error is largest in the center of coarse elements i.e. at large
distances from the coarse model finite element nodes.

\newpage
\subsection{Predictions}
\begin{figure}[h]
\centering
\includegraphics[width=.85\textwidth]{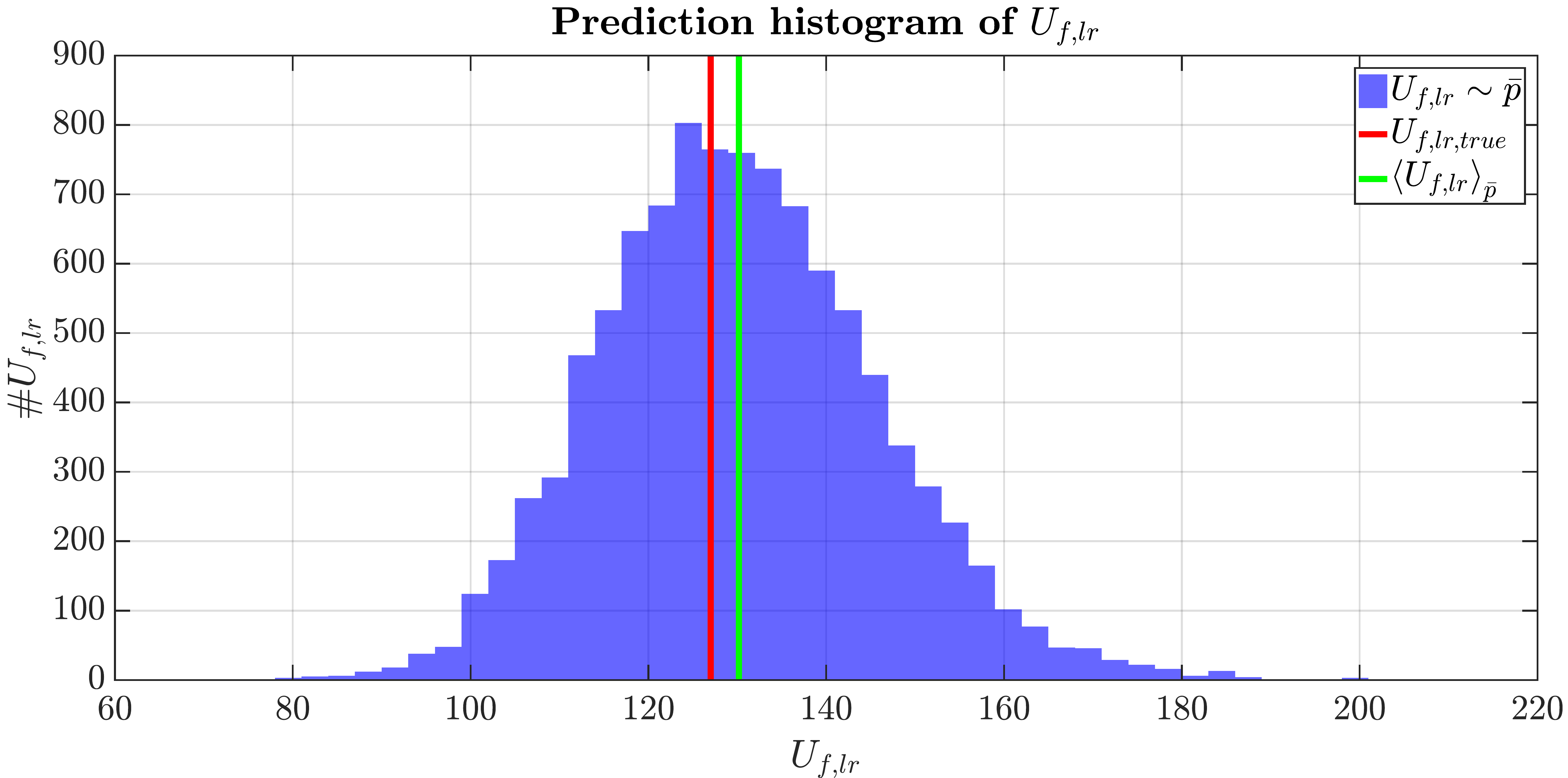}
\caption{Predictive histogram (samples from $\bar{p}(\bs U_{f, lr}|\bs \lambda_f, \bs \theta^*)$) for the temperature
$U_{f,lr}$ of the lower right corner of the domain. The true solution $U_{f,lr,\tm{true}}$ is nicely captured by the
distribution $\bar{p}$.}
\label{predHist}
\end{figure}

\begin{figure}[h]
\centering
\includegraphics[width=.76\textwidth]{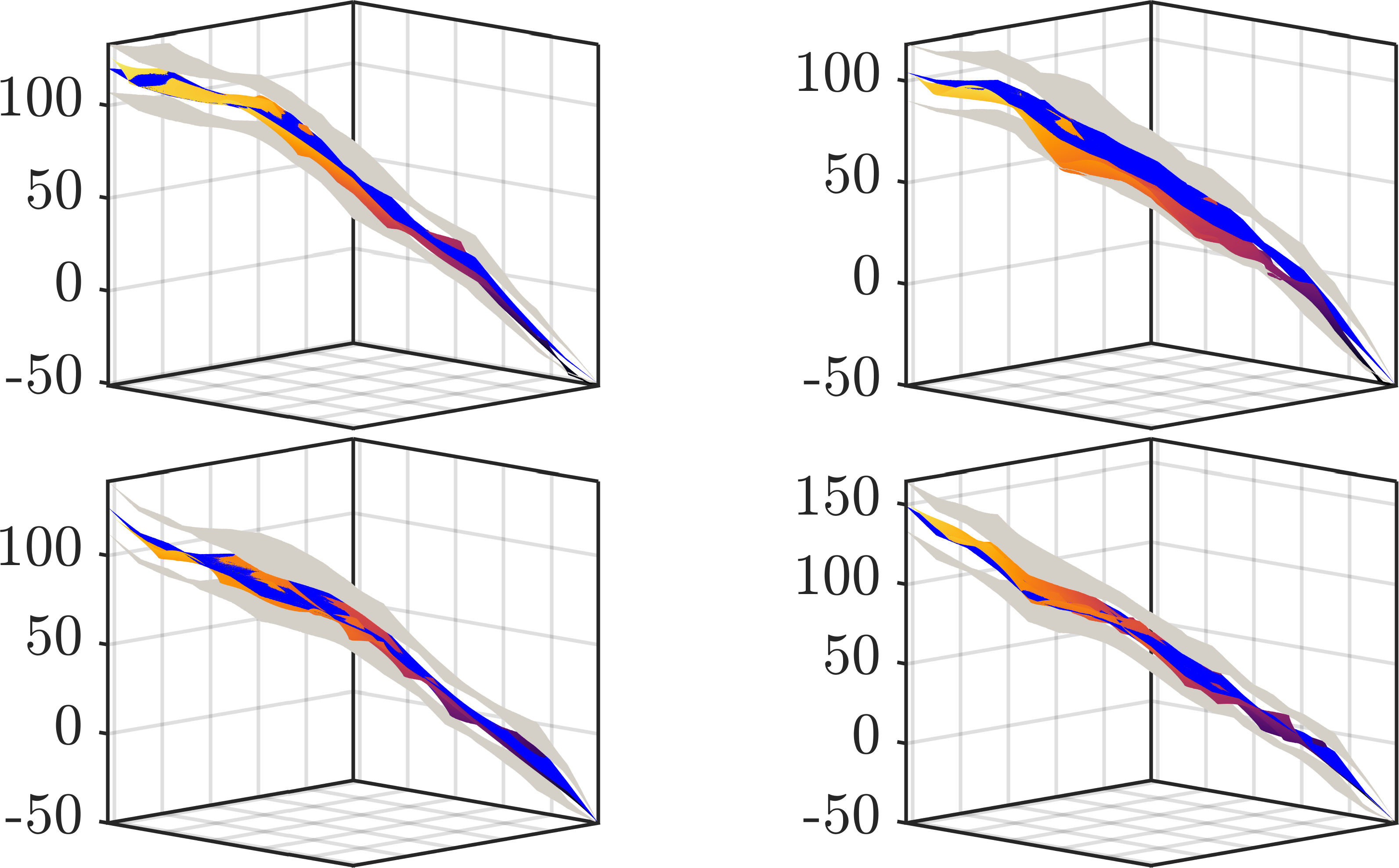}
\caption{Predictions over the whole domain on four different test samples. The true solution (colored) lies in between
$\pm\sigma$ (grey). The predictive mean (blue) is very close to the true solution.}
\label{sandwichMulti}
\end{figure}

As in section \ref{errorData}, for a model with $N_{\tm{train}} = 32$ and $\dim(\bs \lambda_c) = 8\times 8$ and random 
microstructures with  parameter values  $\phi_{\tm{hi}} = 0.2, l = 0.0781, c = 10$, we consider  predictions
by sampling from $\bar{p}(\bs U_f|\bs \lambda_f, \bs \theta_{cf}, \bs \theta_c)$ using 10,000 samples. The predictive
histogram for the temperature $U_{f,lr}$ in the lower right corner of the domain can be seen in Figure \ref{predHist}.
Figure \ref{sandwichMulti} shows a surface plot of the true solution (colored), the predictive mean (blue) $\pm \sigma$
(gray). As can be seen, the true solution $\bs U_f$ is nicely included in $\bar{p}$ everywhere.


\newpage
\section{CONCLUSION}
In this paper, we described a generative Bayesian model which is capable of giving probabilistic predictions to an
expensive fine-grained model (FG) based  only on a small finite number of training data  and multiple solutions of 
a fast,
but less accurate reduced order model (CG). 

In particular, we consider the discretized solution of stochastic PDEs with random coefficients where the FG
corresponds to a fine-scale discretization. Naturally, this comes along with a very high-dimensional vector of input
uncertainties. The proposed model is capable of extracting the most relevant features of those input uncertainties and
gives a mapping to a much lower dimensional space of effective properties (encoding).
These lower dimensional effective properties serve as the input to the CG, which solves the PDE on a much coarser
scale. The last step consists of a probabilistic reconstruction mapping from the coarse- to the fine-scale solution
(decoding).

We demonstrated features and capabilities  of the model proposed for a  $2D$ steady-state 
heat problem, where the fine scale of the conductivity implies (upon discretization) a random input vector of 
dimension $256\times 256$ and the solution of a discretized system of equations of comparable size. In combination 
with a sparsity-enforcing prior, the proposed model
identified the most salient features of the fine-scale conductivity field and allowed accurate predictions of the FG
response using a CGs  of size only $2\times 2, 4\times 4$ and $8\times 8$.
The predictive distribution always included the true FG solution as well as provided uncertainty bounds arising from 
the information loss taking place during the coarse-graining process.

\newpage
\bibliographystyle{abbrv}
\bibliography{library.bib}

\begin{thebibliography}{10}

\bibitem{Bernardo2003}
J.~Bernardo, M.~Bayarri, J.~Berger, A.~Dawid, D.~Heckerman, A.~Smith, and
  M.~West.
\newblock Bayesian factor regression models in the “large p, small n”
  paradigm.
\newblock {\em Bayesian statistics}, 7:733--742, 2003.

\bibitem{Bishop1995}
C.~M. Bishop.
\newblock {\em Neural networks for pattern recognition}.
\newblock Oxford university press, 1995.

\bibitem{Bishop1998}
C.~M. Bishop.
\newblock {\em Latent Variable Models}, pages 371--403.
\newblock Springer Netherlands, Dordrecht, 1998.

\bibitem{Bishop2006}
C.~M. Bishop.
\newblock {\em {Pattern Recognition and Machine Learning}}, volume~4.
\newblock 2006.

\bibitem{Cover2012}
T.~M. Cover and J.~A. Thomas.
\newblock {\em Elements of information theory}.
\newblock John Wiley \& Sons, 2012.

\bibitem{Dempster1977}
A.~P. Dempster, N.~M. Laird, and D.~B. Rubin.
\newblock Maximum likelihood from incomplete data via the em algorithm.
\newblock {\em Journal of the royal statistical society. Series B
  (methodological)}, pages 1--38, 1977.

\bibitem{Figueiredo2003a}
M.~A.~T. Figueiredo.
\newblock {Adaptive sparseness for supervised learning}.
\newblock {\em IEEE Transactions on Pattern Analysis and Machine Intelligence},
  25(9):1150--1159, 2003.

\bibitem{Gahem1991}
R.~G. Gahem, P.~D. Spanos, R.~G. Ghanem, and P.~D. Spanos.
\newblock {\em {Stochastic Finite Elements: A Spectral Approach}}.
\newblock 2003.

\bibitem{Hoffman2013}
M.~D. Hoffman, D.~M. Blei, C.~Wang, and J.~W. Paisley.
\newblock Stochastic variational inference.
\newblock {\em Journal of Machine Learning Research}, 14(1):1303--1347, 2013.

\bibitem{Kennedy2000a}
M.~C. Kennedy and A.~O'Hagan.
\newblock {Predicting the output from a complex computer code when fast
  approximations are available}, 2000.

\bibitem{Koutsourelakis2006}
P.~Koutsourelakis.
\newblock Probabilistic characterization and simulation of multi-phase random
  media.
\newblock {\em Probabilistic Engineering Mechanics}, 21(3):227--234, 2006.

\bibitem{Koutsourelakis2009}
P.-S. Koutsourelakis.
\newblock {Accurate Uncertainty Quantification Using Inaccurate Computational
  Models}.
\newblock {\em SIAM Journal on Scientific Computing}, 31(5):3274--3300, 2009.

\bibitem{Koutsourelakis2005}
P.-S. Koutsourelakis and G.~Deodatis.
\newblock Simulation of binary random fields with applications to two-phase
  random media.
\newblock {\em Journal of Engineering Mechanics}, 131(4):397--412, 2005.

\bibitem{ma_kernel_2011}
X.~Ma and N.~Zabaras.
\newblock Kernel principal component analysis for stochastic input model
  generation.
\newblock {\em Journal of Computational Physics}, 230(19):7311--7331, Aug.
  2011.

\bibitem{MATLAB:2016}
MATLAB.
\newblock {\em version 9.1.0.441655 (R2016b)}.
\newblock The MathWorks Inc., Natick, Massachusetts, 2016.

\bibitem{Perdikaris2015}
P.~Perdikaris, D.~Venturi, J.~O. Royset, and G.~E. Karniadakis.
\newblock {Multi-fidelity modelling via recursive co-kriging and Gaussian –
  Markov random fields Subject Areas :}.
\newblock 2015.

\bibitem{robbins_stochastic_1951}
H.~Robbins and S.~Monro.
\newblock A stochastic approximation method.
\newblock {\em The Annals of Mathematical Statistics}, 22(3):400--407, 1951.

\bibitem{Roberts1995}
A.~Roberts and M.~Teubner.
\newblock Transport properties of heterogeneous materials derived from gaussian
  random fields: bounds and simulation.
\newblock {\em Physical Review E}, 51(5):4141, 1995.

\bibitem{DBLP:journals/jcphy/SchoberlZK17}
M.~Sch{\"{o}}berl, N.~Zabaras, and P.-S. Koutsourelakis.
\newblock Predictive coarse-graining.
\newblock {\em J. Comput. Physics}, 333:49--77, 2017.

\bibitem{Stein1981}
C.~Stein.
\newblock {Stein-1981.pdf}.
\newblock {\em The Annals of statistics}, 9(6):1135--1151, 1981.

\bibitem{Tibshirani1996}
R.~Tibshirani.
\newblock {Royal Statistical Society}.
\newblock {\em Journal of the Royal Statistical Society B}, 58(1):267--288,
  1996.

\bibitem{DBLP:journals/corr/physics-0004057}
N.~Tishby, F.~C.~N. Pereira, and W.~Bialek.
\newblock The information bottleneck method.
\newblock {\em CoRR}, physics/0004057, 2000.

\bibitem{Torquato2001}
S.~Torquato.
\newblock {\em {Random Heterogeneous Materials - Microstructure and Macroscopic
  Properties}}.
\newblock 2001.

\bibitem{Wainwright2008}
M.~J. Wainwright, M.~I. Jordan, et~al.
\newblock Graphical models, exponential families, and variational inference.
\newblock {\em Foundations and Trends{\textregistered} in Machine Learning},
  1(1--2):1--305, 2008.

\bibitem{Weinan2011}
E.~Weinan.
\newblock {\em Principles of multiscale modeling}.
\newblock Cambridge University Press, 2011.

\bibitem{Williams2005a}
C.~E.~R. Williams and C.~K. I.
\newblock {\em {Gaussian Processes for Machine Learning}}.
\newblock 2005.

\bibitem{Zou2007}
H.~Zou, T.~Hastie, and R.~Tibshirani.
\newblock {On the "degrees of freedom" of the lasso}.
\newblock {\em Annals of Statistics}, 35(5):2173--2192, 2007.

\end{thebibliography}

\end{document}